\documentclass[letterpaper]{article} 
\usepackage{aaai2026}  
\usepackage{times}  
\usepackage{helvet}  
\usepackage{courier}  
\usepackage[hyphens]{url}  
\usepackage{graphicx} 
\usepackage{natbib}  
\usepackage{caption} 
\usepackage{algorithm}
\usepackage{algorithmic}
\usepackage{multirow}
\usepackage{booktabs}
\usepackage{newfloat}
\usepackage{listings}
\DeclareCaptionStyle{ruled}{labelfont=normalfont,labelsep=colon,strut=off} 
\floatstyle{ruled}
\newfloat{listing}{tb}{lst}{}
\floatname{listing}{Listing}
\title{Bridging Simulation and Reality: Cross-Domain Transfer with Semantic 2D Gaussian Splatting}

\author{
    Jian Tang\textsuperscript{\rm 1}\equalcontrib,
    Pu Pang\textsuperscript{\rm 1,2}\equalcontrib,
    Haowen Sun\textsuperscript{\rm 1},
    Chengzhong Ma\textsuperscript{\rm 1},
    Xingyu Chen\textsuperscript{\rm 1},
    Hua Huang\textsuperscript{\rm 2},
    Xuguang Lan\textsuperscript{\rm 1}\thanks{Corresponding author}
}

\begin{document}
\maketitle

\begin{abstract}
Cross-domain transfer in robotic manipulation remains a longstanding challenge due to the significant domain gap between simulated and real-world environments. Existing methods such as domain randomization, adaptation, and sim-real calibration often require extensive tuning or fail to generalize to unseen scenarios. To address this issue, we observe that if domain-invariant features are utilized during policy training in simulation, and the same features can be extracted and provided as the input to policy during real-world deployment, the domain gap can be effectively bridged, leading to significantly improved policy generalization. Accordingly, we propose Semantic 2D Gaussian Splatting (S2GS), a novel representation method that extracts object-centric, domain-invariant spatial features. S2GS constructs multi-view 2D semantic fields and projects them into a unified 3D space via feature-level Gaussian splatting. A semantic filtering mechanism removes irrelevant background content, ensuring clean and consistent inputs for policy learning. To evaluate the effectiveness of S2GS, we adopt Diffusion Policy as the downstream learning algorithm and conduct experiments in the ManiSkill simulation environment, followed by real-world deployment. Results demonstrate that S2GS significantly improves sim-to-real transferability, maintaining high and stable task performance in real-world scenarios.

\end{abstract}


\section{Introduction}
Sim-to-real cross-domain transfer remains a fundamental challenge in robotic learning, as policies trained purely in simulation often fail to generalize to real-world environments due to the discrepancies in visual appearance, object diversity, and environmental complexity~\cite{tobin2017domainrandomization}. To address these challenges, a wide range of approaches has been explored, including domain randomization, domain adaptation,and sim-real calibration or translation. 
Domain randomization introduces variability in textures, lighting, and dynamics during training to improve robustness, whereas domain adaptation learns mappings between simulated and real domains~\cite{peng2018sim2real,ganin2016domainadversarial,yang2025novel}. Sim-real calibration approaches~\cite{schperberg2023real,chukwurah2024sim2real} focus on calibrating simulators by matching sensor, material, and dynamics properties of the real world, effectively minimizing the discrepancy at training time.

\begin{figure}[t]
\centering
\includegraphics[width=0.9\columnwidth]{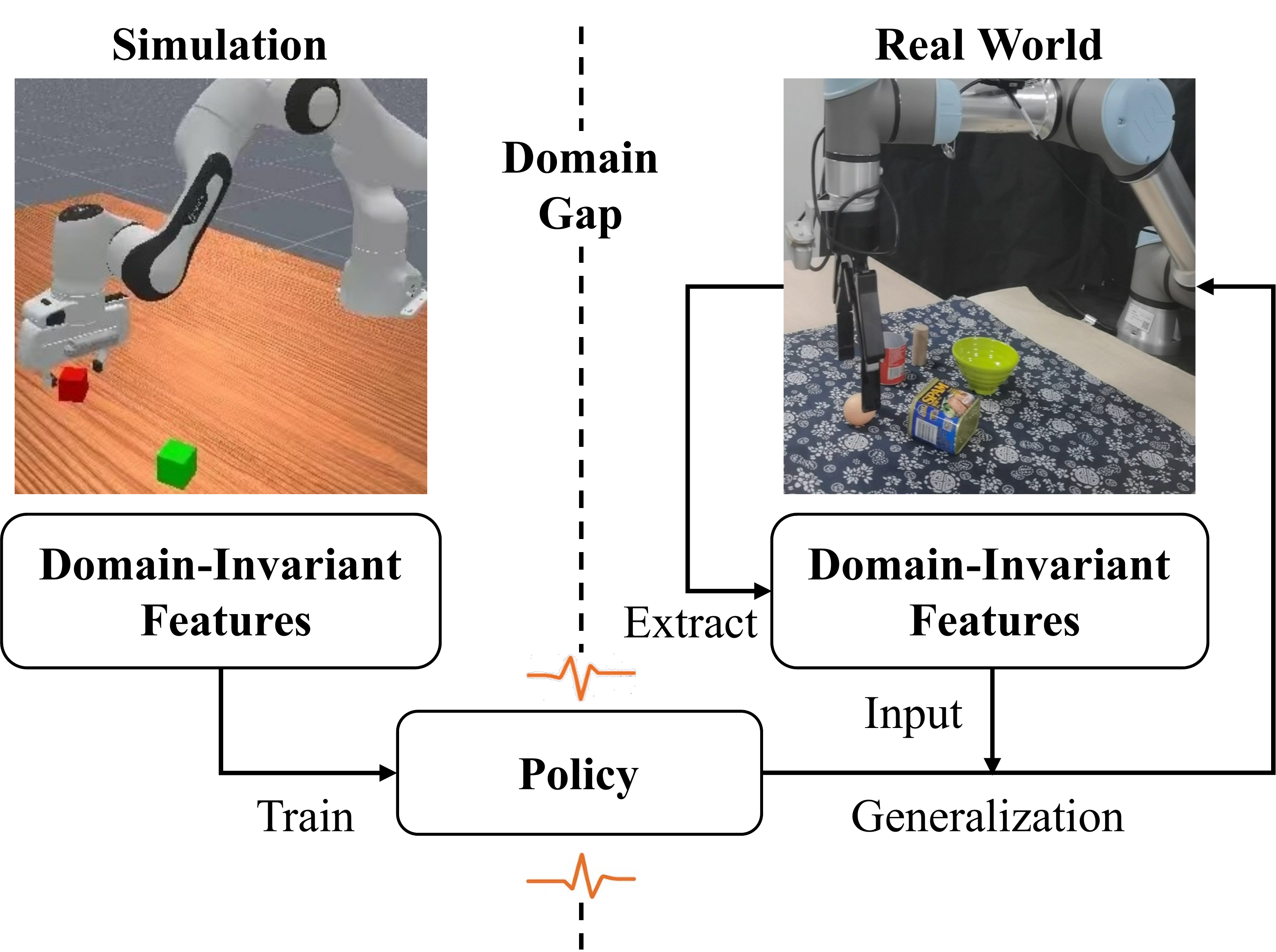} 
\caption{Sim-to-real domain transfer with domain-invariant features. Although there exist significant domain gap between the simulation and the real world, both share a common feature space of domain-invariant features. Leveraging these domain-invariant features to train the policy in simulation, and directly extracting the same type of features from real-world observations as the policy input, can mitigate the domain gap and enhance sim-to-real generalization.}
\label{fig1}
\end{figure}
However, each approach suffers from inherent limitations. Domain randomization and adaptation frequently require extensive hyperparameter tuning and large amounts of diverse data~\cite{saito2021tune,hansen2020self}. Meanwhile, sim-real calibration methods~\cite{valassakis2020crossing,huang2023went}demand considerable engineering effort and still struggle to generalize across the long-tail of real-world variations not captured in simulation tuning.

In the area of representation learning, researchers have explored various methods to reduce the sim-to-real gap. Early approaches relied on RGB-based neural networks to learn visual features, aiming to achieve cross-domain alignment in an end-to-end manner~\cite{zhang2018visual}. However, such methods struggle to maintain semantic and spatial consistency across multiple viewpoints, which limits the generalization of policies in complex real-world scenes. Subsequent work introduced point cloud representations, explicitly encoding 3D spatial structures to enhance geometric reasoning~\cite{qi2017pointnet}. Yet, point clouds are typically sparse and unstructured, making it difficult to embed high-level semantic features. More recently, NeRF-based methods have emerged~\cite{mildenhall2021nerf}, offering a convenient way to incorporate semantic information while preserving multi-view spatial consistency. Nevertheless, their heavy computational cost and slow rendering speed make them impractical for real-time robotic control. Overall, these approaches still struggle to efficiently and accurately extract domain-invariant spatial features that are essential for robust sim-to-real transfer.

\begin{figure*}[t]
\centering
\includegraphics[width=1.0\textwidth]{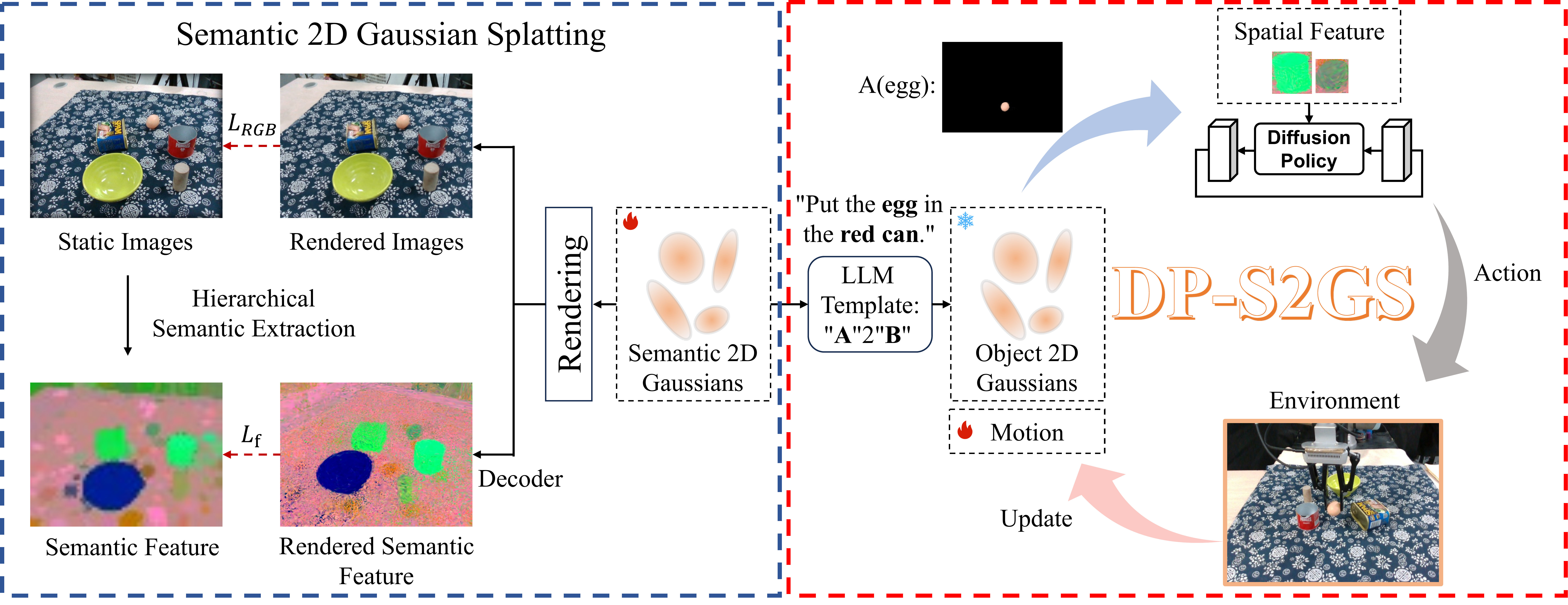} 
\caption{S2GS Overview. S2GS aims to extract domain-invariant spatial features to support robust cross-domain policy transfer. In the initial stage, S2GS extracts hierarchical semantic features of multi-view images and optimizes the semantic 2D Gaussian Splatting field and the decoder. While executing, a semantic retrieval module queries and filters task-relevant objects while removing background distractions. The resulting domain-invariant spatial features serve as compact and clean inputs for downstream diffusion policy learning. After manipulation, S2GS supports dynamic scene updating to maintain accurate scene representation in real-time, satisfying the 
requirements of online robotic control.}
\label{fig2}
\end{figure*}

From the perspective of cognitive neuroscience, humans can still perform a learned basic manipulation skill effectively even when facing unfamiliar environments. This cross-domain generalization ability arises from the formation of neuronal selectivity, which combines long-term experiential accumulation with the training effects of neural plasticity, enabling the brain to extract stable and invariant visual representations.~\cite{storm2024integrative} have shown that neurons in the human medial temporal lobe selectively activate in response to images of faces, animals, objects, or scenes, refining complex visual input into essential features and storing them in memory. Likewise, in robotic policy learning for manipulation primitives (e.g., grasping, pushing, and stacking), if domain-invariant spatial features (such as object centroids, surface normals, and orientations) are employed during simulation training, and the same features are provided as input during deployment in real-world settings, which can significantly mitigates the domain gap and enhances sim-to-real generalization.

Motivated by this insight, we aim to extract domain-invariant spatial features at the representation level, thereby effectively reducing the domain gap in cross-domain policy transfer.Therefore, we propose \textbf{Semantic 2D Gaussian Splatting (S2GS)}. This approach can accurately extract domain-invariant spatial features of task-relevant objects, enhancing the reliability of policies in cross-domain transfer. At the same time, S2GS offers high editability, allowing flexible removal of irrelevant environmental distractions so that the policy focuses on task-critical targets. In addition, it provides strong real-time performance, meeting the requirements of online robotic control in real-world environments. Specifically, S2GS takes multi-view images as input, constructs a 2D feature representation for each frame, and maps these features into a unified 3D semantic space through feature-level Gaussian Splatting. Subsequently, S2GS applies a semantic retrieval mechanism to remove redundant background information, retaining only the semantic features of key target objects. From these, it extracts domain-invariant spatial features of task-relevant objects, providing a clean and focused input for policy learning and effectively reducing environmental distractions.

To validate the effectiveness of S2GS in sim-to-real transfer, we integrate it with Diffusion Policy~\cite{chi2023diffusion} as the downstream learning module and conduct experiments on a suite of tasks in the ManiSkill2~\cite{gu2023maniskill2} simulation platform. Subsequently, the learned policies are deployed in real-world robotic environments. Experimental results demonstrate that S2GS significantly improves transfer robustness and maintains strong task performance in the real world, outperforming baseline methods that rely on traditional visual inputs.

Our contributions can be summarized as follows:
\begin{itemize}
    \item We introduce S2GS, a novel representation method that extracts object-centric, domain-invariant spatial features, effectively reducing the domain gap in sim-to-real policy transfer.
    \item S2GS provides high editability and real-time capability, enabling flexible removal of task-irrelevant background distractions while maintaining performance suitable for online robotic control in real-world environments.
    \item We combine S2GS with Diffusion Policy to validate its effectiveness in cross-domain sim-to-real transfer.
\end{itemize}

\section{Related Work}

\noindent\textbf{Cross-Domain Sim-to-Real Transfer}
Transferring policies from simulation to the real world remains a long-standing challenge in robotic learning due to the domain gap in visual appearance, object diversity, and environmental complexity. Policies trained solely on simulated data often fail when confronted with unseen objects or cluttered backgrounds.  

One widely adopted approach is domain randomization~\cite{tobin2017domain,muratore2022robot}, which exposes agents to randomized textures, lighting, and dynamics in simulation so that they can generalize to real-world variations at test time. While simple and effective in some settings, domain randomization usually requires large-scale randomized environments and can produce overly conservative policies~\cite{tiboni2023domain}.  

Another line of research focuses on domain randomization and domain adaptation to narrow the appearance gap between simulated and real data. Methods such as image translation using GraspGAN~\cite{bousmalis2018using} and randomized-to-canonical adaptation networks (RCAN)~\cite{james2019sim} translate simulated images into a canonical real-world style. More recent approaches~\cite{peng2018sim} extend this idea by aligning both perception and dynamics distributions.  

Beyond image-level adaptation, representation learning techniques aim to build domain-invariant state encodings. 
Latent feature learning~\cite{hansen2020self}, semantic segmentation~\cite{kirillov2023segment}, and structured state abstractions~\cite{sadeghi2017cad2rl} have been shown to improve robustness to domain shifts by removing task-irrelevant variations in raw observations. Diffusion-based visuomotor policies~\cite{chi2023diffusionpolicy} and language-conditioned policies~\cite{ahn2022do} further leverage richer inputs to enhance cross-domain generalization, but they still rely heavily on dense RGB observations, which remain sensitive to clutter and appearance variation.  

Sim-real calibration aims to align simulation with the real world by tuning sensor, material, and dynamics parameters, but it often requires considerable engineering effort and still fails to generalize to unseen real-world variations not captured during tuning~\cite{peng2018sim2real,qureshi2024splatsim}.

However, these approaches still struggle to provide domain-invariant spatial features. They either rely heavily on dense RGB observations, which are sensitive to clutter and appearance variations, or require extensive domain randomization and image-level adaptation. In contrast, S2GS extracts clean, object-centric and domain-invariant spatial features and filters out task-irrelevant background information, effectively reducing the domain gap 

\noindent\textbf{3D Representations for Robotic Manipulation.}
The ability to perceive and model 3D scenes is fundamental for robotic manipulation. Neural Radiance Fields (NeRF)~\cite{mildenhall2021nerf} significantly advanced this capability by creating high-fidelity implicit 3D representations from sparse images. In robotics, this was quickly leveraged for manipulation tasks. Early works demonstrated that NeRF could reconstruct object geometry to generate grasp poses~\cite{ichnowski2021dex}, laying the groundwork for more complex interactions. Subsequent research further explored using NeRF-derived representations to learn dense object descriptors for manipulation~\cite{yen2022nerf}.

However, a primary limitation of these early NeRF-based manipulation approaches is their reliance on a time-consuming, per-scene optimization process. This often requires dense multi-view inputs and lengthy training for each new object or scene, making them impractical for dynamic, real-time robotic applications where speed is critical. While generalizable NeRFs~\cite{dai2022graspnerf} aimed to alleviate this by enabling zero-shot reconstruction, they still inherit the computational overhead of querying an implicit model.

More recently, 3D Gaussian Splatting (3DGS)~\cite{kerbl3Dgaussians} has emerged as a powerful alternative that directly addresses the speed bottleneck. By representing scenes explicitly as a collection of 3D Gaussians, 3DGS allows for real-time, high-fidelity rendering and geometric querying. This makes it highly suitable for robotics, leading to immediate applications in manipulation. These include real-time grasp detection~\cite{10607869}, the creation of photorealistic simulations to improve sim-to-real transfer~\cite{qureshi2024splatsim}, and the modeling of dynamic 3DGS in simulation for robot motion estimation~\cite{lu2024manigaussian, chai2025gaf}. Nevertheless, these prior works often presuppose that the reconstructed simulation must maintain high fidelity to the real world. This emphasis on static accuracy tends to overlook the dynamic, closed-loop interactions between the physical environment, the digital 3D model, and the robot's actions, which is critical for robust manipulation. 
Our work mainly focus on the interaction between digital model and the real world, and how to use the semantic 2D Gaussians to bridge the gap between the simulation and physical worlds, improving the generalization of robot manipulation from simulation to the real world.

\section{Methods}
In this section, we present our method for building a semantic 2D Gaussian Splatting representation (see Figure~\ref{fig2} for an overview). We begin with a preliminary overview of Gaussian Splatting techniques (Section~\ref{sec:preliminary}). We then describe the core components of our approach: (1) a hierarchical semantic feature lifting mechanism for S2GS (Section~\ref{sec:semantic_2dgs}); (2) an open-vocabulary spatial feature extraction module for task target object (Section~\ref{sec:querying}); and (3) a real-time optimization strategy for dynamic scene updates (Section~\ref{sec:updating}).

\subsection{Preliminary}\label{sec:preliminary}
\subsubsection{3D Gaussian Splatting.}
3DGS~\cite{kerbl3Dgaussians} represents 3D scenes using a set of learnable 3D Gaussians \(\{\mathcal{G}_i\}\). Each Gaussian \(\mathcal{G}_i\) is parameterized by its center position \(\mathbf{\mu}_i\), rotation quaternion \(\mathbf{q}_i\), scaling vector \(\mathbf{s}_i\), color \(c_i\), and opacity \(o_i\). The spatial distribution of the Gaussian is defined by:
\begin{equation}
\mathcal{G}_i(\mathbf{x} | \mathbf{\mu}_i, \mathbf{\Sigma}_i) = e^{-\frac{1}{2}(\mathbf{x} - \mathbf{\mu}_i)^T \mathbf{\Sigma}_i^{-1} (\mathbf{x} - \mathbf{\mu}_i)},
\end{equation}
where the covariance matrix \(\mathbf{\Sigma}_i\) = \(R_i S_i S_i^T R_i^T\), derived from the quaternion \(\mathbf{q}_i\) and the scaling vector \(\mathbf{s}_i\).

This representation enables real-time rendering via a differentiable rasterization pipeline. For a given camera view with world-to-camera matrix \(\mathbf{W}\) and intrinsics \(\mathbf{K}\), the 3D center \(\mathbf{\mu}_i\) is projected to its 2D pixel coordinate \(\mathbf{\mu'}_i\) = \(\mathbf{K} \mathbf{W} [\mathbf{\mu}_i, 1]^T\). Concurrently, the 3D covariance matrix \(\mathbf{\Sigma}_i\) is projected into 2D space:
\begin{equation}
\mathbf{\Sigma'}_i = J \mathbf{W} \mathbf{\Sigma}_i \mathbf{W}^T J^T,
\end{equation}
where \(J\) is the Jacobian of the affine approximation of the projective 
transformation.

The final color \(C\) for a pixel at coordinate \(\mathbf{u}\) is computed via \(\alpha\)-blending, accumulating the colors of all overlapping Gaussians sorted from front to back:
\begin{equation}
C(\mathbf{u}) = \sum_{i \in N}  T_i \alpha_i c_i,
T_i = \prod_{j=1}^{i-1} (1 - \alpha_j),
\end{equation}
where \(c_i\) is the view-dependent color  represented by spherical harmonics (SH), and \(\alpha_i\) is the product of the opacity \(o_i\) and the value of the projected 3D Gaussian at the pixel coordinate:
\begin{equation}
\alpha_i = o_i \mathcal{G'}_i(\mathbf{u} | \mathbf{\mu'}_i, \mathbf{\Sigma'}_i).
\end{equation}

\subsubsection{2D Gaussian Splatting.}
While 3DGS excels at novel view synthesis, it faces challenges in inconsistency depth when rendered from different viewpoints, leading to noisy or flawed geometry. 

To address these limitations, 2D Gaussian Splatting (2DGS)~\cite{huang20242d} represents scenes using a collection of 2D Gaussian primitives. Unlike the 3D counterparts, these primitives are defined as oriented elliptical disks in 3D space, characterized by a 2D scaling vector instead of a 3D one. This allows them to better align with object surfaces. 
When rendering, 2DGS employs an explicit ray-splat intersection~\cite{weyrich2007hardware} to determine how each primitive projects onto the image plane. The final pixel color is then computed using a similar \(\alpha\)-blending process.

\subsection{Semantic 2DGS}\label{sec:semantic_2dgs}
\subsubsection{Appearance modeling.}
Following 2DGS, we adopt a combination of an L1 loss and a D-SSIM term between the rendered image \(\hat{I}\) and the input image \(I\):
\begin{equation}
\mathcal{L}_{rgb} = (1 - \lambda)\mathcal{L}_{1}(\hat{I}, I) + \lambda\mathcal{L}_{D-SSIM}(\hat{I}, I),
\end{equation}
where \(\lambda\) is set to 0.2.

\subsubsection{Hierarchical Semantic Extraction.}
To capture semantics at multiple granularities, we distill hierarchical semantic features by CLIP~\cite{radford2021learning} and SAM~\cite{kirillov2023segment}, following~\cite{qin2024langsplat, ji2024graspsplats}. 
Specifically, we first feed the input image to CLIP and SAM to obtain the the semantic feature \(\mathbf{f}^I_{raw}\) of total image and masks \(\{\mathbf{M}^I\}\) of different regions. Then, we get the global semantic feature by averaging the semantic vectors of the pixels in each region to ensure the consistency of the same object:
\begin{equation}
\mathbf{f}^I_{g}(\mathbf{u}) = \frac{1}{|\mathbf{M}^I(\mathbf{u})|} \sum \mathbf{f}^I_{raw}(\mathbf{u}) \odot \mathbf{M}^I(\mathbf{u}),
\end{equation}
where \(\mathbf{M}^I(\mathbf{u})\) is the mask which pixel \(\mathbf{u}\) belongs to.
After that, we crop the image according to the masks and feed them to CLIP to get the object semantic feature \(\mathbf{f}^I_{obj}\). Finally, we remap the object semantic feature into original image space to get the local semantic feature \(\mathbf{f}^I_{l}\).

\subsubsection{Semantic Feature Rendering.}
Inspired by recent works~\cite{qin2024langsplat, zhou2024feature, chen2024pgsr} that distill semantic features from 2D vision foundation models into 3DGS, we augment each 2D Gaussian primitive \(\mathcal{G}_i\) with a learnable, high-dimensional semantic feature vector \(\mathbf{f}_i\). 
The semantic feature is rendered by the \(\alpha\)-blending process:
\begin{equation}
\mathbf{f}(\mathbf{u}) = \sum_{i \in N} T_i \alpha_i \mathbf{f}_i.
\end{equation}

\subsubsection{Semantic Modeling.}
To maintain rendering efficiency, each 2D Gaussian primitive stores a compact, low-dimensional semantic feature. Since these features have lower dimensionality than the raw hierarchical features, we introduce a shallow MLP decoder \(\Psi\) to map the rendered features \(\mathbf{\hat{f}}\) back to the original high-dimensional space. This decoder predicts both global and local semantic features, which are supervised using a cosine similarity loss against their respective ground-truth counterparts:
\begin{equation}
\mathbf{\hat{f}}_{l}, \mathbf{\hat{f}}_{g} = \Psi(\mathbf{\hat{f}}),
\end{equation}
\begin{equation}
\mathcal{L}_{f} = \mathbf{sim}(\mathbf{\hat{f}}_{g}, \mathbf{f}^I_{g}) + \lambda_{l}\mathbf{sim}(\mathbf{\hat{f}}_{l}, \mathbf{f}^I_{l}),
\end{equation}
where \(\mathbf{f}^I_{l}\) and \(\mathbf{f}^I_{g}\) are the precomputed local and global semantic features, respectively, \(\mathbf{sim}(f_1, f_2)\) = \(1 - \frac{f_1 \cdot f_2}{\|f_1\| \|f_2\|}\) is the cosine similarity function, and \(\lambda_{l}\) = 0.5 is the weight for the local semantic distillation loss.

\subsubsection{Geometry Regularization.}
Encouraged by~\cite{qi2018geonet, huang20242d, jiang2024gaussianshader, long2024adaptive, chen2024pgsr}, we also apply local consistency of normal and depth:
\begin{equation}
\mathcal{L}_{n} = \sum_{\mathbf{u}} \omega(\mathbf{u})\left(1 - N(\mathbf{u}) \cdot N_d(\mathbf{u})\right),
\end{equation}
where \(N(\mathbf{u})\), \(N_d(\mathbf{u})\), and \(\omega(\mathbf{u})\) is the normal, the depth normal, and the rendering weight of the pixel \(\mathbf{u}\), respectively, computed as:
\begin{equation}
    N(\mathbf{u}) = \sum_{i \in N} T_i \alpha_i n_i,
\end{equation}
\begin{equation}
    D(\mathbf{u}) = \sum_{i \in N} T_i \alpha_i d_i,
\end{equation}
\begin{equation}
    N_d(\mathbf{u}) = {\nabla_x D(\mathbf{u}) \times \nabla_y D(\mathbf{u}) \over \left\|{\nabla_x D(\mathbf{u}) \times \nabla_y D(\mathbf{u})}\right\|},
\end{equation}
\begin{equation}
    \omega(\mathbf{u}) = \sum_{i \in N} T_i \alpha_i,
\end{equation}
where \(d_i\) and \(n_i\) are the depth and normal of the \(i\)-th 2D Gaussian primitive, respectively.

Our total loss function is a weighted sum of an appearance loss, a semantic distillation loss, and the geometric regularization losses:
\begin{equation}
\mathcal{L} = \mathcal{L}_{rgb} + \lambda_{f} \mathcal{L}_{f} + \lambda_{n} \mathcal{L}_{n},
\end{equation}
where \(\lambda_{f}\) = 0.1 and \(\lambda_{n}\) = 0.05 are the weights for the semantic distillation loss and the geometric regularization loss, respectively.

\subsection{Task target object spatial feature extraction.}\label{sec:querying}

Once the semantic 2DGS field is constructed, we can query the object by natural language with CLIP. By computing the similarity of decoded 2D Gaussian primitive's semantic feature and the query language, we can get the initial object Gaussian primitives. 
However, the initial primitives may contain noise and lack spatial consistency with the underlying object structure. To address these issues, we employ a two-stage refinement process to improve segmentation quality and extract precise object poses.

\noindent\textbf{Instance Clustering.} We first apply DBSCAN clustering to filter out noise primitives. The clustering operates on the positions of initial object Gaussian primitives with parameters \(\epsilon = 0.02\)m and minimum samples of 15, effectively eliminating spatially isolated noise while preserving coherent object regions.

\noindent\textbf{Geometric Completion.} We select the largest cluster as the target object and apply convex hull completion to fill potential holes in the segmentation. This addresses incomplete coverage caused by insufficient semantic similarity scores in uniform surfaces or occluded regions, ensuring complete object representation by including all primitives within the convex hull.

Finally, we derive a compact spatial feature of the target object by aggregating its Gaussian primitives into a centroid representation, which provides a concise and stable state input for the diffusion policy.

In our decision stage, the diffusion policy~\cite{ho2020denoising,chi2023diffusion,ze20243d} is conditioned on the object-centric spatial features $s$ extracted by S2GS and the current robot state(end-effector pose) $q$. Starting from a sequence of random noisy actions, the policy gradually denoises them into a feasible action sequence through the reverse diffusion process, formulated as:
\begin{equation}
a_{k-1} = \alpha_k 
\Big(
a_k - \gamma_k \, \epsilon_\theta(a_k, k, s, q)
\Big) 
+ \sigma_k \mathcal{N}(0, I),
\end{equation}
where $a_k$ denotes the noisy action at step $k$, $\epsilon_\theta$ is the conditional denoising network, $\alpha_k,\gamma_k,\sigma_k$ are controlled by the noise scheduler, and $\mathcal{N}(0,I)$ is standard Gaussian noise.

The training objective adopts a mean squared error (MSE) formulation to predict the noise added at step $k$ in the forward diffusion process:
\begin{equation}
\mathcal{L} = 
\mathrm{MSE}\Big(
\epsilon^k,\;
\epsilon_\theta\big(
\sqrt{\bar{\alpha}_k}\, a_0
+ \sqrt{\bar{\beta}_k}\, \epsilon^k,\;
k,\, s,\, q
\big)
\Big),
\end{equation}
where $a_0$ is the ground-truth expert action from the dataset, $\epsilon^k$ is the Gaussian noise sampled at step $k$, and $\bar{\alpha}_k$ and $\bar{\beta}_k$ are the noise scheduling parameters.

\subsection{Dynamic Scene Updating}\label{sec:updating}


\begin{figure}[t]
\centering
\includegraphics[width=1.0\columnwidth]{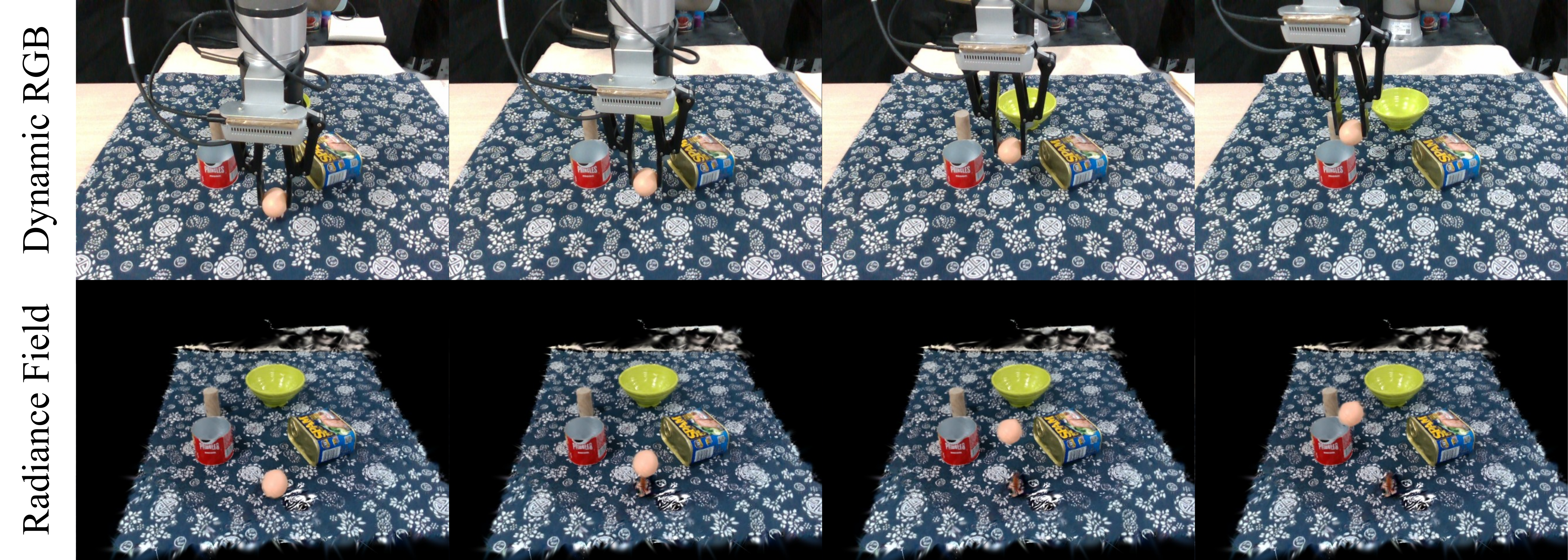}
\caption{Dynamic scene updating process. Our method tracks object motion during manipulation tasks by optimizing SE(3) transformations to maintain accurate scene representation in real-time.}
\label{fig:dynamic_tracking}
\end{figure}

During manipulation tasks, a static 2DGS representation, constructed from initial multi-view captures, quickly becomes outdated as the robot interacts with objects. While periodic re-scanning can update the scene, this process is too time-consuming for real-time robotic applications. Existing dynamic update methods also present significant challenges. Splat-MOVER~\cite{shorinwa2024splat} require a predefined motion function, and GraspSplats~\cite{ji2024graspsplats} depends on complex tracking pipelines that leverage multi-view cameras and depth sensors.

To overcome these limitations, we directly optimize the object's motion. Assuming rigidity, this motion is parameterized by an SE(3) transformation, represented by a 3D translation vector and a 4D rotation quaternion. We solve for these motion parameters by minimizing the L1 loss between the rendered and input images with the mask of the motion object. To further accurate the optimization, we incorporate the known robotic manipulation as a strong motion prior. The dynamic updating process is shown in Figure~\ref{fig:dynamic_tracking}.

\section{Experiments}

\begin{figure*}[t]
\centering
\includegraphics[width=1.0\textwidth]{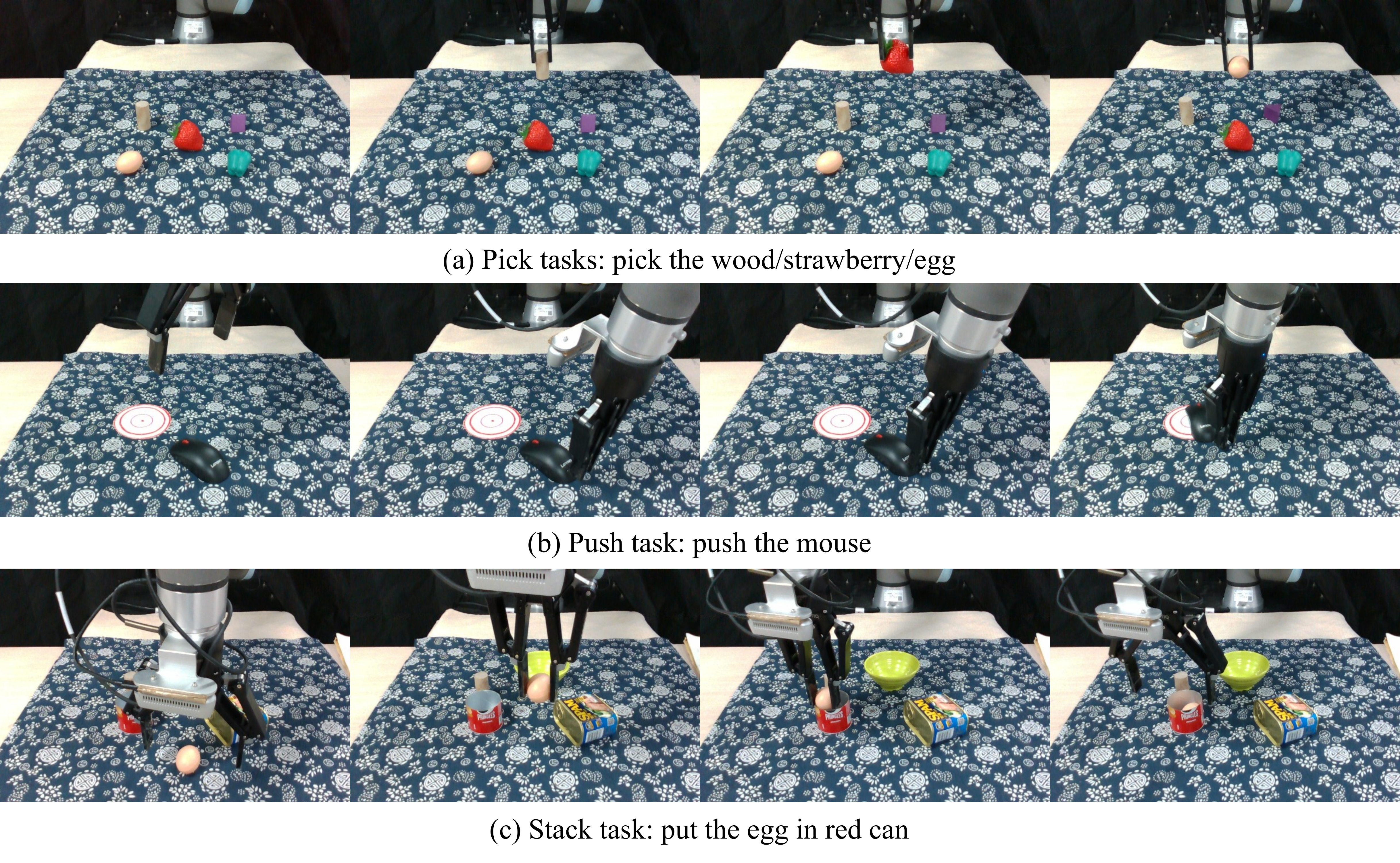}
\caption{Real-world results. Our method achieves high success rates in real-world manipulation tasks, demonstrating the effectiveness of our S2GS representation.}
\label{fig:real_world}
\end{figure*}
    
We train our diffusion policy in simulation and directly perform cross-domain transfer to real-world environments to evaluate its effectiveness for sim-to-real robotic manipulation. This section investigates the following key questions:

\begin{itemize}
\item How accurately can S2GS extract the spatial features of target objects?  
\item Does incorporating S2GS-derived spatial features significantly improve the model’s sim-to-real transfer performance?  
\end{itemize}

\subsection{Experiment Setup}
In simulation, the model is trained on the ManiSkill2~\cite{gu2023maniskill2} using a 7-DOF Franka Panda robot. Since our ultimate goal is to deploy the model on a real-world 6-DOF UR5 robot, we remove the joint angle information from the state inputs to prevent overfitting to robot-specific kinematics and avoid transfer failure due to the degree-of-freedom mismatch.

In the real-world evaluation, the trained policy is deployed on a UR5 robotic arm. An Intel RealSense D435i camera mounted on the robot gripper performs an initial panoramic scan of the workspace to construct the semantic 2DGS field, which we only optimize 7,000 steps for efficiency. A second fixed RealSense D435i camera continuously updates the 2DGS representation, ensuring that it remains synchronized with dynamic changes in the environment.

\subsection{Baseline}
\begin{figure}[h]
\centering
\includegraphics[width=1.0\columnwidth]{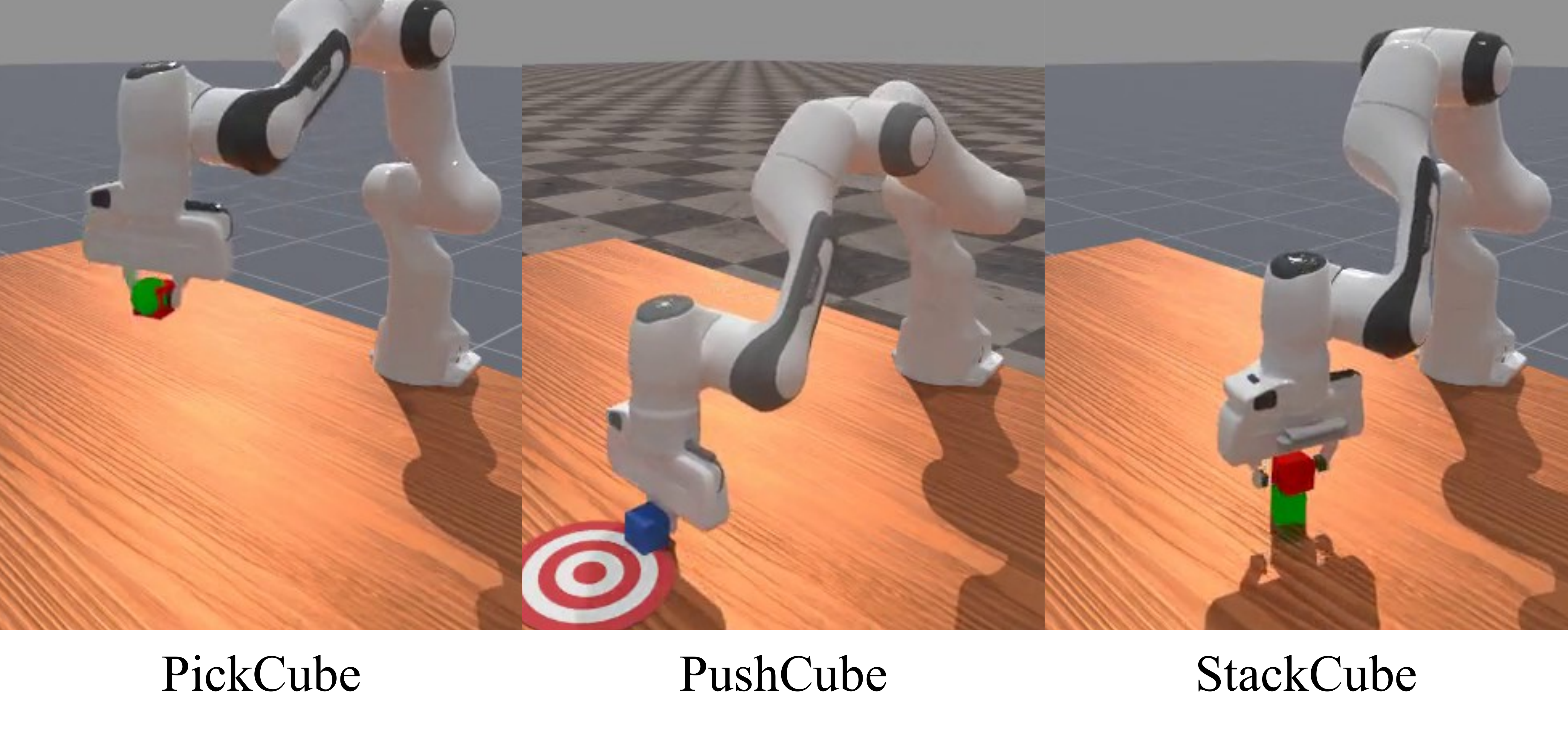}
\caption{Three tasks in simulation: PickCube, PushCube, and StackCube.}
\label{fig:simtask}
\end{figure}

We use Diffusion Policy from the ManiSkill Imitation Learning benchmark as our baseline. The three tasks we evaluate are PickCube, PushCube, and StackCube, as shown in Figure~\ref{fig:simtask}. For each task and input modality (state or RGB), we select 100 trajectories and train the model for 30,000 steps, reporting the best score as the final result.

\subsection{Simulation Result}

\begin{table}[]
\centering
\small
\begingroup
\begin{tabular}{l|c|ccc}
\hline
\multirow{2}{*}{Input} & \multirow{2}{*}{Method} & \multicolumn{3}{c}{Task} \\
\cline{3-5}
& & Pick & Push & Stack \\
\hline
\multirow{4}{*}{RGB} & DP-\(RGB\) & 0.76 & 0.89 & 0.04 \\
    & DP-\(RGB_{+GS}\) & 0.91 & 0.92 & 0.29 \\
    & DP-\(RGB_{+GS}+S2GS\) & \textbf{1} & \textbf{1} & 0.94 \\
    & DP-\(S2GS\) & \textbf{1} & \textbf{1} & \textbf{0.97} \\
\hline
State & DP-\(State\) & \textbf{1} & \textbf{1} & \textbf{0.97} \\
\hline
\end{tabular}
\endgroup
\caption{Success rate for different methods on three tasks. The performance comparison demonstrates the effectiveness of our approach across different manipulation tasks in simulation environment.}
\label{tab:success_rate}
\end{table}

\begin{table*}[t]
\centering
\begin{tabular}{ll ll ll}
\toprule
\textbf{Pick Task} & \textbf{Success} & \textbf{Push Task} & \textbf{Success} & \textbf{Stack Task} & \textbf{Success} \\
\midrule
Pick the wood & 5/5 & Push the mouse & 3/5 & Put the egg in red can & 4/5 \\
Pick the strawberry & 3/5 & Push the can & 5/5 & Put the egg in bowl & 5/5 \\
Pick the egg & 5/5 & Push the bucket & 5/5 & Put the wood on can & 3/5 \\
\midrule
\textbf{Average} & \textbf{86.7\%} & ~ & \textbf{86.7\%} & ~ & \textbf{80.0\%} \\
\bottomrule
\end{tabular}
\caption{Success rate (\%) of DP-S2GS on real-world pick, push, and stack tasks. Our method achieves consistently high performance across fundamental manipulation primitives in real-world scenarios.}
\label{tab:task_success}
\end{table*}

We evaluate our method on PickCube, PushCube, and StackCube tasks in the ManiSkill simulation environment. The results, summarized in Table~\ref{tab:success_rate}, demonstrate the effectiveness of our S2GS representation across different input modalities.

\noindent\textbf{RGB-based Methods.} Raw RGB inputs (DP-RGB) achieve reasonable performance on Pick (0.76) and Push (0.89) tasks but fail dramatically on StackCube (0.04), highlighting the challenge of multi-object reasoning. Adding Gaussian rendering (DP-\(RGB_{+GS}\)) provides substantial improvements (0.91, 0.92, 0.29), demonstrating the value of filtering background distractions. Our semantic enhancement (DP-\(RGB_{+GS}+S2GS\)) further boosts performance to perfect scores on Pick and Push tasks and 0.94 on StackCube.

Crucially, we explore abandoning the traditional visual encoder paradigm by using purely S2GS-derived spatial features (DP-\(S2GS\)). This achieves the best performance (1.0, 1.0, 0.97), revealing that raw RGB can introduce interference and that S2GS provides cleaner, task-relevant features.

\noindent\textbf{State-based Upper Bound.} Our DP-S2GS method matches the ground-truth state performance (DP-State: 1.0, 1.0, 0.97), demonstrating that S2GS extracts spatial features functionally equivalent to ground-truth states, enabling robust sim-to-real transfer.

\subsection{Real-world Result}
To evaluate the sim-to-real generalization of our approach, we directly deploy the trained policy onto UR5 robotic arm without any additional fine-tuning. The real-world experiments cover three fundamental manipulation tasks—Pick, Push, and Stack—on a variety of unseen household objects that differ significantly from those in the simulation environment in terms of shape, material, and appearance.

As shown in Table~\ref{tab:task_success}, S2GS achieves consistently high success rates across all task types. For pick tasks, which are relatively simpler, the average success rate reaches 86.7\%. A slight drop is observed in the "Pick the strawberry" subtask due to the difficulty in achieving a force-closure grasp—caused by the object's irregular surface and the limited contact geometry of the two-finger gripper. In push tasks, the overall success rate is also 86.7\%, with lower performance in the "Push the mouse" subtask. This can be attributed to the mouse's partially symmetric geometry, which results in uneven torque response when pushed along non-symmetric directions. Stack tasks pose the greatest challenge, with an average success rate of 80.0\%. The subtask "Put the wood on can" exhibits a notable drop in performance, particularly when the can is placed upright, increasing the vertical placement difficulty and introducing a domain gap that exceeds the model's adaptation capacity. These results demonstrate S2GS enable robust cross-domain transfer in complex real-world scenarios.

\subsection{Ablation}

\begin{table}[h]
\centering
\small
\begin{tabular}{l|ccc}
\hline
Representation & PSNR $\uparrow$ & SSIM $\uparrow$ & LPIPS $\downarrow$ \\
\hline
3DGS & 21.22 & 0.741 & 0.310 \\
S2GS (Ours) & \textbf{23.68} & \textbf{0.861} & \textbf{0.202} \\
\hline
\end{tabular}
\caption{Appearance quality comparison between 3DGS and S2GS in the real-world scene. S2GS consistently outperforms 3DGS across all evaluation metrics, demonstrating superior rendering quality and reduced perceptual artifacts in real-world robotic manipulation scenarios.}
\label{tab:ablation_s2gs_vs_3dgs}
\end{table}

\begin{figure}[h]
\centering
\includegraphics[width=1.0\columnwidth]{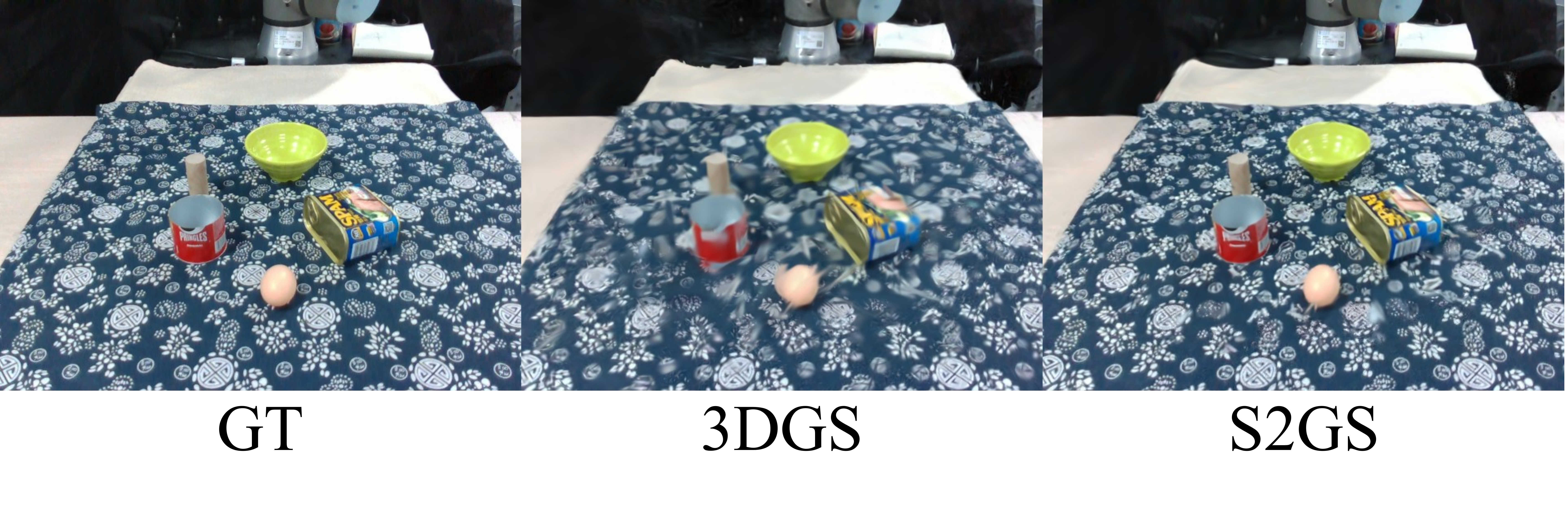}
\caption{Visual comparison of the rendered image from 3DGS and S2GS model. S2GS produces visually sharper and more detailed reconstructions with better object boundary definition and reduced artifacts compared to the traditional 3DGS approach.}
\label{fig:ablation_visuals}
\end{figure}    
We conduct an ablation study to validate the effectiveness of S2GS over the 3D one. 
We optimize the Gaussian model for 7,000 steps, and then evaluate the appearance quality of the rendered image.
As shown in Table~\ref{tab:ablation_s2gs_vs_3dgs} and Figure~\ref{fig:ablation_visuals}, S2GS achieves a PSNR improvement of 2.46 dB (from 21.22 to 23.68) and SSIM enhancement of 0.12 (from 0.741 to 0.861), while reducing LPIPS by 0.108 (from 0.310 to 0.202). This improvement is attributed to the 2D Gaussian primitives' superior ability to align with object surfaces, reducing the geometric inconsistencies that typically arise in 3D approaches when rendered from different viewpoints.

\subsection{Conclusion and Limitations}
We propose Semantic 2D Gaussian Splatting (S2GS), a novel representation that extracts domain-invariant spatial features to address the domain gap in sim-to-real robotic manipulation. Integrated with Diffusion Policy, S2GS achieves robust generalization across simulation and real-world tasks, and improves sim-to-real transfer.

However, S2GS has not yet incorporated other domain-invariant spatial features such as surface normals, which may hinder generalization in contact-sensitive tasks. We will address this in future work.




\bibliography{aaai2026}
\newpage
\appendix

\section*{Appendix}

\subsection*{A. Simulator Training Details}
\begin{figure}[h]
\centering
\includegraphics[width=1.0\columnwidth]{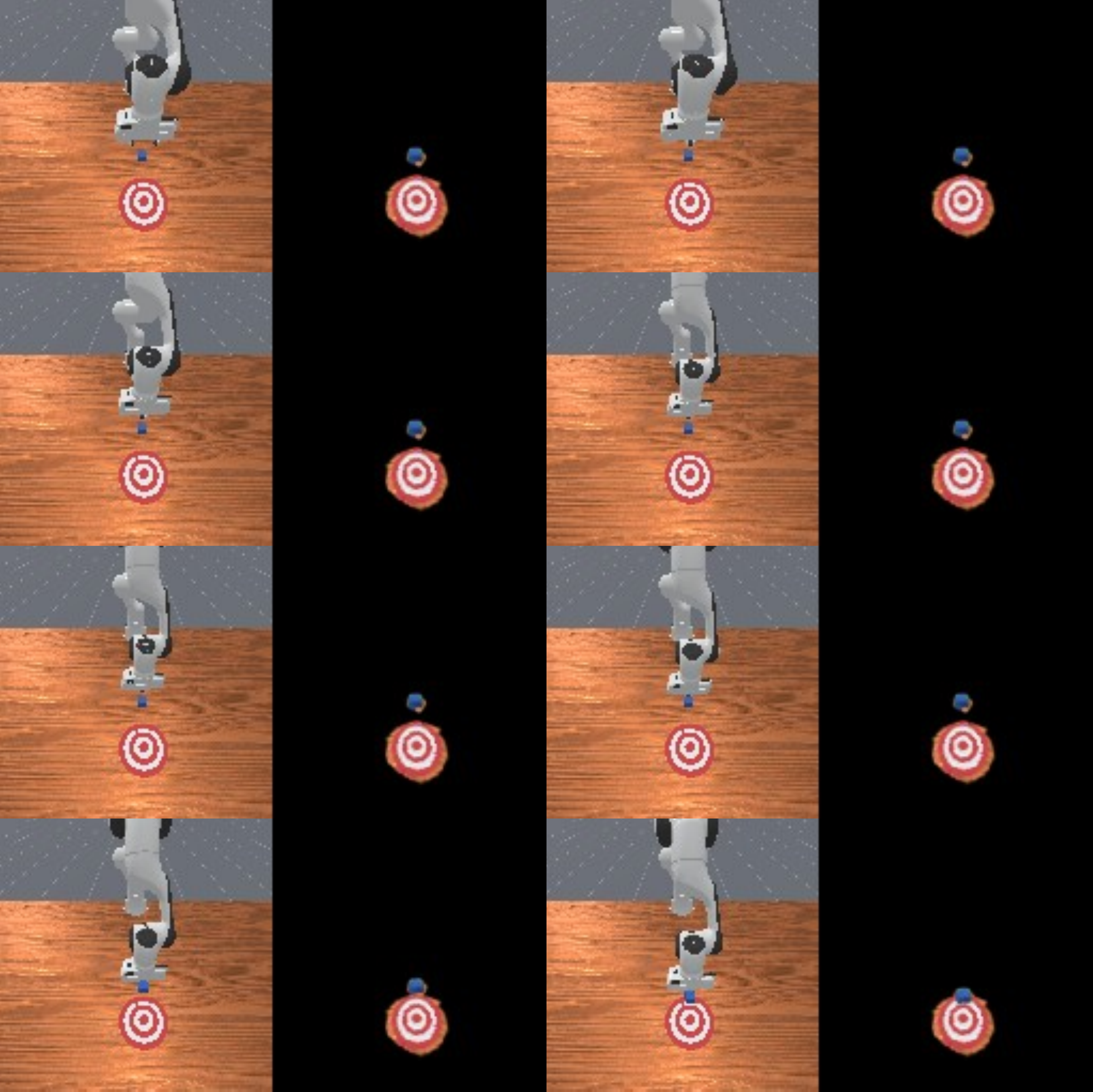}
\caption{Visualization of the policy with S2GS representation during the evaluation process in the ManiSkill2 simulation environment.}
\label{fig:sim_eval}
\end{figure} 
In simulation, we use the ManiSkill2~\cite{gu2023maniskill2} benchmark environment, featuring a 7-DOF Franka Panda robotic arm. Figure~\ref{fig:sim_eval} showing the S2GS model during the evaluation process.

\subsection*{B. Experiment and Implementation Details}

For real-world deployment, we adopt a 6-DOF UR5 robotic arm mounted with an Intel RealSense D435i RGB-D camera on the gripper. Additionally, a second fixed D435i camera is used to continuously update the 2DGS representation throughout the task execution. 

To construct the S2GS representation, we use the gripper-mounted camera to capture 39 RGB images while rotating 360° around the scene. Additionally, one panoramic RGB image is captured from the fixed camera, resulting in a total of 40 images used to generate the initial semantic field. This representation is then used to extract object-centric spatial features and provide clean, structured inputs for policy learning. The S2GS training is performed for 7,000 steps using an NVIDIA RTX 3090 GPU, taking approximately 7 minutes to complete. With a more powerful A800 GPU, the optimization time can be reduced to around 3 minutes.

\begin{figure}[h]
\centering
\includegraphics[width=1.0\columnwidth]{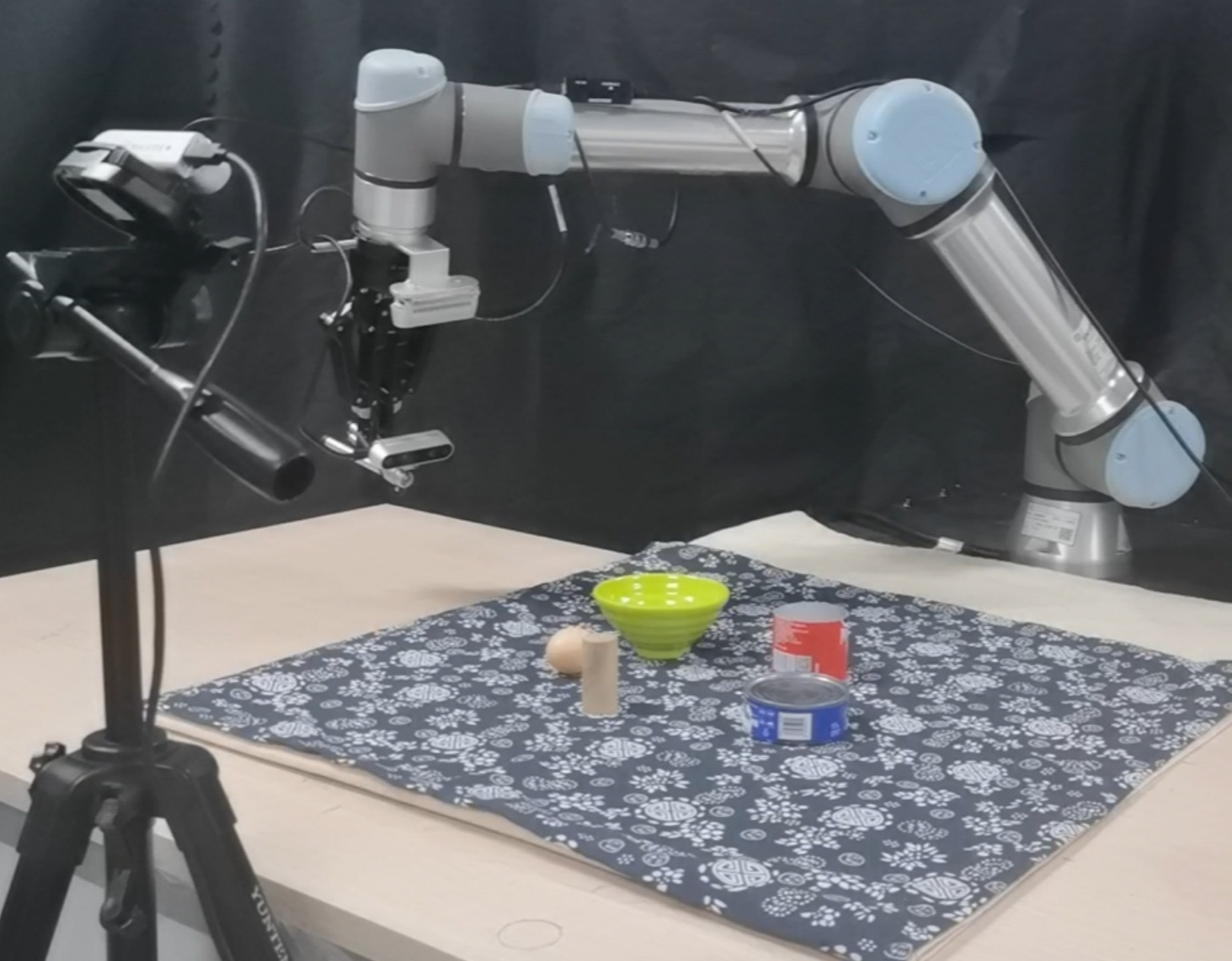}
\caption{Real-World Construction Details of S2GS.}
\label{fig:experiment_scan_scene}
\end{figure}

\subsection*{C. Real-World Experiment Analysis}

\begin{figure*}[h]
\centering
\includegraphics[width=0.92\textwidth]{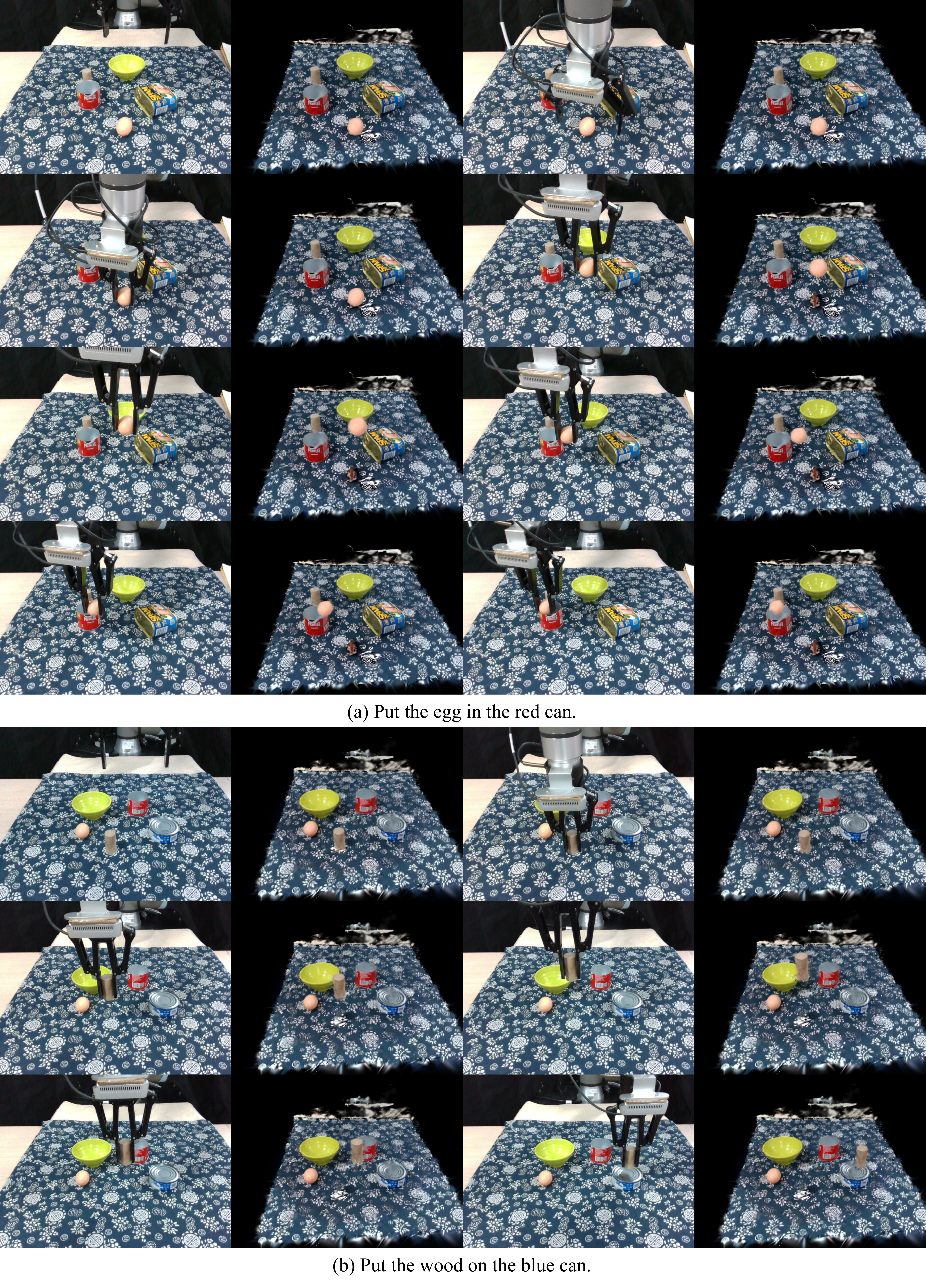}
\caption{Tracking process visualization during object manipulation. Our method maintains accurate position estimation throughout the manipulation sequence, demonstrating real-time updates of the semantic 2DGS representation.}
\label{fig:tracking_process}
\end{figure*}

\begin{figure*}[h]
\centering
\includegraphics[width=1.0\textwidth]{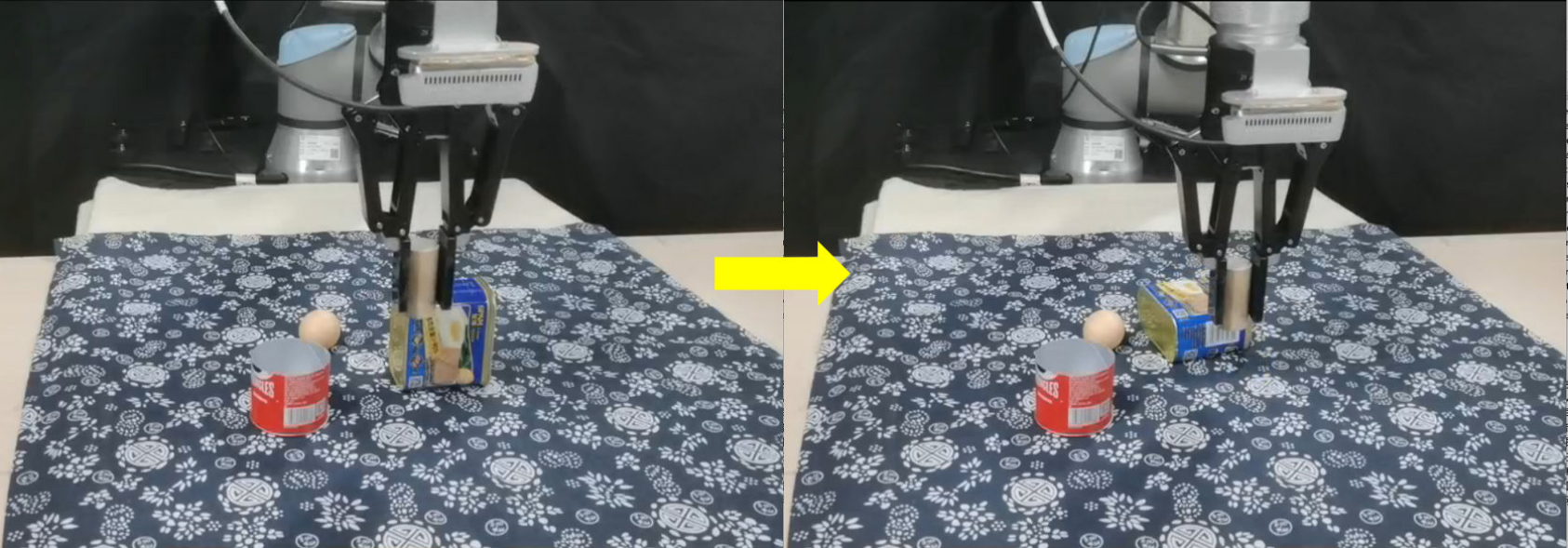}
\caption{Failure case. When the blue can is too tall, the gripper may collide with it during placement, causing it to tip over.}
\label{fig:failure_case}
\end{figure*}

We evaluate the policy on three real-world manipulation tasks: PickCube, PushCube, and StackCube, consistent with the simulation setup. For each task, we run 15 real-world trials to assess success rate and qualitative performance. The results show that:

\begin{itemize}
\item Our method achieves stable success rates across all tasks without any additional domain adaptation.
\item The S2GS-derived spatial features enable the diffusion policy to operate effectively despite visual and physical discrepancies between simulation and reality.
\item In challenging scenes with background clutter or partial occlusion, S2GS demonstrates robustness by filtering out irrelevant features and maintaining accurate object localization.
\end{itemize}

Figure~\ref{fig:tracking_process} visualizes the tracking process during manipulation. This dynamic scene updating is crucial for maintaining accurate representations for robotic manipulation tasks.

During manipulation, our method achieves real-time tracking with translation errors at the millimeter level. The lightweight optimization process adds minimal computational overhead to the overall pipeline. The semantic filtering mechanism effectively reduces the computational burden by focusing updates only on task-relevant regions. With the interaction with robot, we optimize the motion for only 10 steps every frame after the action, suitable for robotic control. 

Figure~\ref{fig:failure_case}shows the failure case. For example, in the Pick the strawberry task, the irregular shape of the strawberry makes it difficult for the two-finger gripper to achieve a force-closure grasp from certain angles, resulting in grasp failures.In the Put the wood on the blue can task, if the blue can is too tall, the gripper may collide with it during placement, causing it to fall over.This is mainly because the training objects in simulation are relatively short cubes, and such tall objects in the real world fall far beyond the generalization range of the trained model, leading to failure.

Overall, the real-world experiments confirm that our proposed S2GS representation enables strong cross-domain generalization, significantly enhancing the practicality of simulation-trained robotic policies.

\end{document}